\documentclass[11pt,a4paper]{article}
\usepackage[hyperref]{acl2021}
\usepackage{times}
\usepackage{latexsym}

\usepackage{microtype}

\usepackage{microtype}
\usepackage{booktabs}
\usepackage{amsmath}
\usepackage{flexisym}
\usepackage{dcolumn}
\usepackage{kotex}
\usepackage{hhline}
\usepackage{verbatim}
\usepackage{multirow}
\usepackage{amssymb}
\usepackage{rotating}
\usepackage{subfigure}
\usepackage{dblfloatfix}
\usepackage{enumitem}
\usepackage{framed}
\usepackage[utf8]{inputenc}
\usepackage[english]{babel}
\usepackage{xcolor}
\usepackage{tabulary}
\usepackage{color}
\usepackage{arydshln}

\newcolumntype{L}[1]{>{\raggedright\let\newline\\\arraybackslash\hspace{0pt}}m{#1}}
\newcolumntype{C}[1]{>{\centering\let\newline\\\arraybackslash\hspace{0pt}}m{#1}}
\newcolumntype{R}[1]{>{\raggedleft\let\newline\\\arraybackslash\hspace{0pt}}m{#1}}

\aclfinalcopy 

\title{UMIC: An Unreferenced Metric for Image Captioning via Contrastive Learning}

\author{Hwanhee Lee$^{1}$, Seunghyun Yoon$^{2}$, Franck Dernoncourt$^{2}$ \\ \bf Trung Bui$^{2}$~\and Kyomin Jung$^{1}$\\
$^{1}$Dept. of Electrical and Computer Engineering, Seoul National University, Seoul, Korea \\
$^{2}$Adobe Research, San Jose, CA, USA,\\
{\tt \{wanted1007, kjung\}@snu.ac.kr} \\
{\tt \{syoon, franck.dernoncourt, bui\}@adobe.com}\\
}

\date{}

\begin{document}
\maketitle
\begin{abstract}
Despite the success of various text generation metrics such as BERTScore, it is still difficult to evaluate the image captions without enough reference captions due to the diversity of the descriptions. In this paper, we introduce a new metric UMIC, an \textbf{U}nreferenced \textbf{M}etric for \textbf{I}mage \textbf{C}aptioning which does not require reference captions to evaluate image captions. Based on Vision-and-Language BERT, we train UMIC to discriminate negative captions via contrastive learning.
Also, we observe critical problems of the previous benchmark dataset (i.e.,  human annotations) on image captioning metric, and introduce a new collection of human annotations on the generated captions. We validate UMIC on four datasets, including our new dataset, and show that UMIC has a higher correlation than all previous metrics that require multiple references. We release the benchmark dataset and pre-trained models to compute the UMIC\footnote{https://github.com/hwanheelee1993/UMIC}.
\end{abstract}

\section{Introduction}
\begin{figure}[!t]
\small
\centering
\includegraphics[width=0.95\columnwidth]{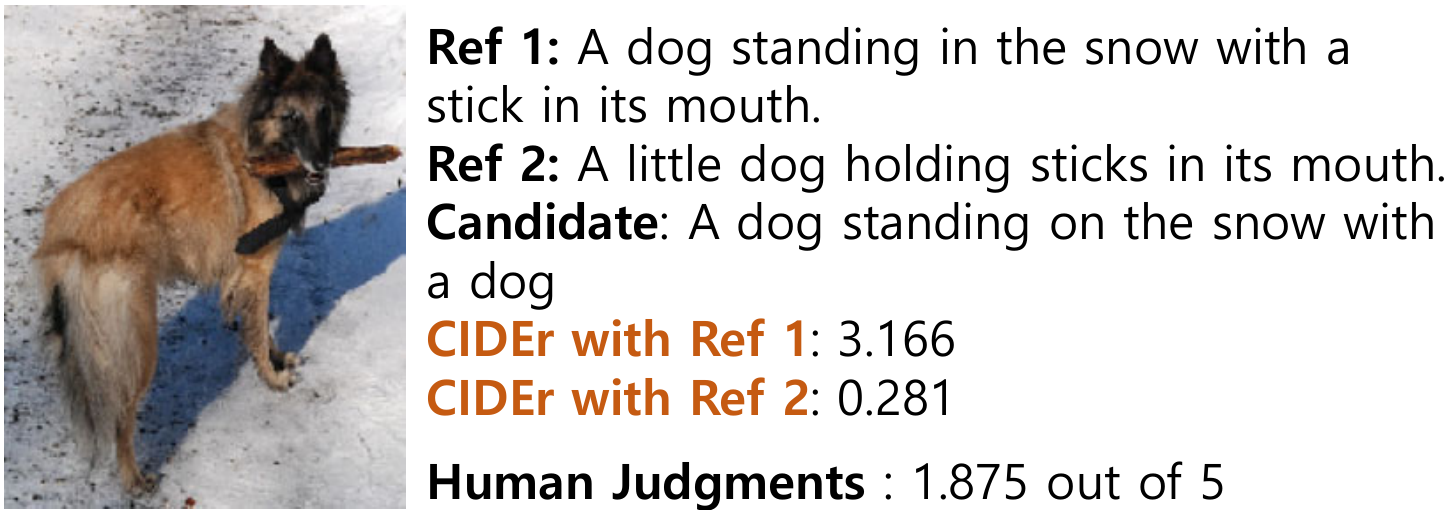}
\caption{
An example where the metric score for a given candidate caption varies significantly depending on the reference type.
}
\vspace{-4mm}
\label{fig_intro}
\end{figure}

Image captioning is a task that aims to generate a description that explains the given image in a natural language.
While there have been many advances for caption generation algorithms~\cite{vinyals2015show, anderson2018bottom} and target datasets~\cite{fang2015captions, sharma2018conceptual}, few studies~\cite{vedantam2015cider, anderson2016spice, cui2018learning, lee2020vilbertscore} have focused on assessing the quality of the generated captions. Especially, most of the evaluation metrics only use reference captions to evaluate the caption although the main context is an image.
However, as shown in Figure~\ref{fig_intro}, since there are many possible reference captions for a single image, a candidate caption can receive completely different scores depending on the type of reference~\cite{yi2020improving}.
Because of this diverse nature of image captions, reference-based metrics usually use multiple references which are difficult to obtain. To overcome this limitation, we propose UMIC, an \textbf{U}nreference \textbf{M}etric for \textbf{I}mage \textbf{C}aptioning, which is not dependent on the reference captions and use an image-caption pair to evaluate a caption. 
We develop UMIC upon UNITER~\cite{chen2020uniter} which is a state-of-the-arts pre-trained representation for vision-and-language tasks.
Since UNITER is pre-trained to predict the alignment for large amounts of image-text pairs, we consider that UNITER can be a strong baseline for developing an unreferenced metric.
We fine-tune UNITER via contrastive learning, where the model is trained to compare and discriminate the ground-truth captions and diverse synthetic negative samples.
We carefully prepare the negative samples that can represent most of the undesirable cases in captioning, such as \textit{grammatically incorrect}, \textit{irrelevant to the image}, or \textit{relevant but have wrong keyword}.
\newline
\indent When evaluating the metric's performance, it is required to compare the correlations between human judgments and the metric's evaluation score for given datasets. We choose three standard benchmark datasets (i.e., Composite~\cite{aditya2015images}, Flickr8k~\cite{hodosh2013framing}, PASCAL-50s~\cite{vedantam2015cider}) and further analyze the quality of the dataset. Interestingly, we found that there exist critical issues in the benchmark datasets, such as poor-label or polarized-label. To perform a rigorous evaluation as well as stimulate the research in this area, we collect new 1,000 human judgments for the model-generated caption. Finally, we evaluate our proposed metric on four benchmark datasets, including our new dataset. Experimental results show that our proposed unreferenced metric is highly correlated with human judgments than all of the previous metrics that use reference captions.

\section{Related Work} 
\paragraph{Image Captioning Metrics}
Following other text generation tasks such as dialogue systems and machine translation, n-gram similarity metrics such as BLEU~\cite{papineni2002bleu}, ROUGE~\cite{lin2004rouge} and METEOR~\cite{banerjee2005meteor} are widely used to evaluate an image caption. Especially, CIDEr~\cite{vedantam2015cider}, which weights each n-gram using TF-IDF, is widely used. SPICE~\cite{anderson2016spice} is a captioning metric based on scene graph. BERTScore~\cite{zhang2019bertscore}, which computes the similarity of the contextualized embeddings, are also used. BERT-TBR~\cite{yi2020improving} focuses on the variance in multiple hypothesis and ViLBERTScore (VBTScore) ~\cite{lee2020vilbertscore} utilizes ViLBERT~\cite{lu2019vilbert} to improve BERTScore. \\
\indent Different from these metrics, VIFIDEL~\cite{madhyastha2019vifidel} computes the word mover distance~\cite{kusner2015word} between the object labels in the image and the candidate captions, and it does not require reference captions. Similar to VIFIDEL, our proposed UMIC does not utilize the reference captions. However, UMIC directly uses image features and evaluates a caption in various perspectives compared to VIFIDEL.

\paragraph{Quality Estimation}
Quality Estimation (QE) is a task that estimates the quality of the generated text without
using the human references and this task is same as developing an unreferenced metric. QE is widely established in machine translation (MT) tasks~\cite{specia2013quest, martins2017pushing, specia2018findings}. Recently, \cite{levinboim-etal-2021-quality} introduces a large scale human ratings on image-caption pairs for training QE models in image captioning tasks. Our work also trains caption QE model, (i.e. unreferenced captioning metric) but we do not use human ratings to train the metric. Instead, we create diverse synthetic negative samples and train the metric with these samples via ranking loss.
\begin{figure}[t]
\small
\centering
\includegraphics[width=0.95\columnwidth]{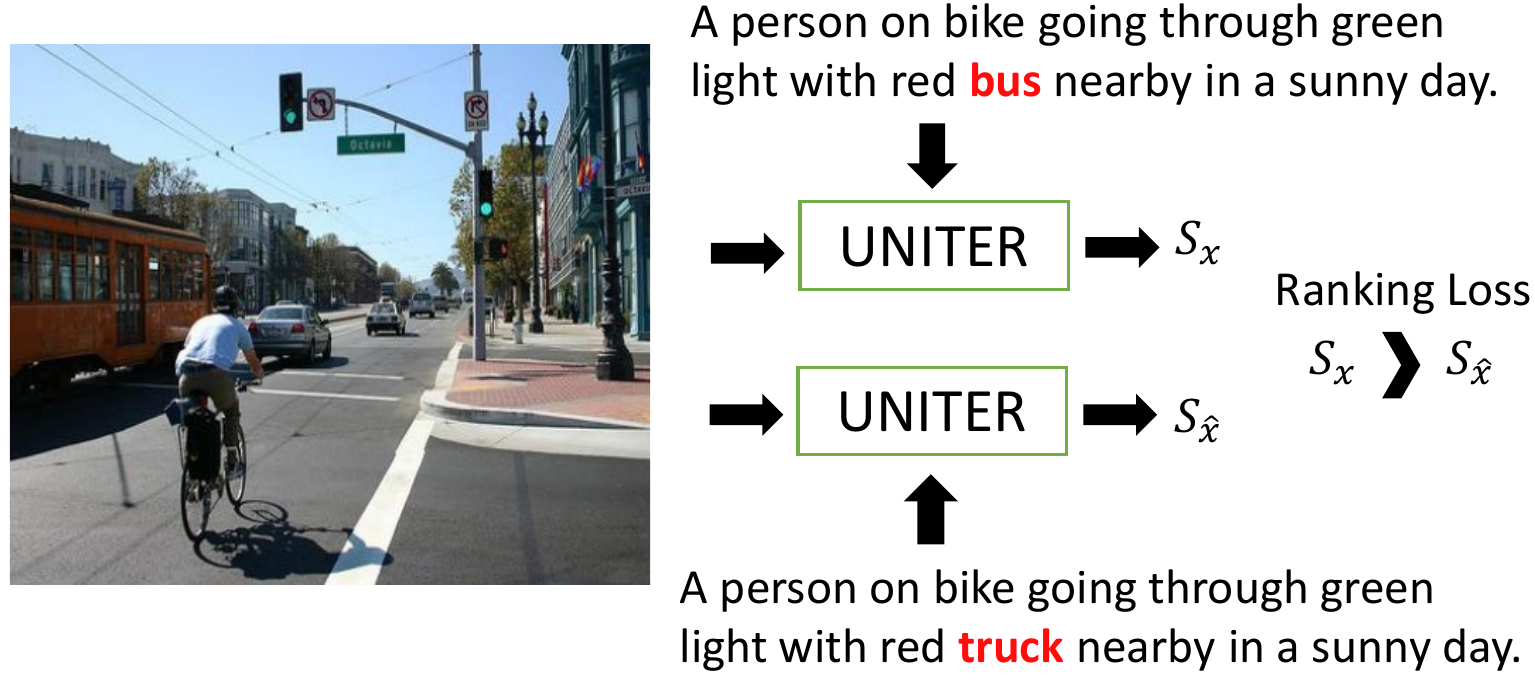}
\caption{
Overall training procedure of UMIC. Given an image $I$, a positive caption $x$ and a negative caption $\hat{x}$, we compute the score of each image-caption pair $S_x$ and $S_{\hat{x}}$ using UNITER respectively. Then, we fine-tune UNITER using raking loss that $S_x$ is higher than $S_{\hat{x}}$.
}
\vspace{-3mm}
\label{fig_overall}
\end{figure}

\section{UMIC}
We propose UMIC, an unreferenced metric for image captioning using UNITER. We construct negative captions using the reference captions through the pre-defined rules. Then, we fine-tune UNITER to distinguish the reference captions and these synthetic negative captions to develop UMIC.

\subsection{Modeling}
Since UNITER is pre-trained to predict the alignment of large amounts of image-text pairs, we use the output of the layer that predicts this alignment as the baseline of UMIC to be fine-tuned. 
Specifically, we compute the score of a caption $\text{S}(I,X)$ for given image $I=({i}_1, ..., {i}_N)$ and $X=({x}_1, ..., {x}_T)$ as follows.
\newline
\indent We first compute the contextual embedding for $I$ and $X$ using UNITER to get the joint representation of image and text as follows. 
\begin{equation}
i_{[CLS]}, i_{1}, ... , i_{N}, x_{1}, ..., x_{T} = \scriptstyle{\text{UNITER}(I, X)},
\end{equation} 
where $i_{[CLS]}$ is a joint representation of the input image and input caption. Then we feed it into a single fully-connected layer to get a score as follows.
\begin{equation}
\text{S}(I,X) = sigmoid(Wi_{[CLS]} + b), 
\end{equation}
where $W$ and $b$ are trainable parameters.

\subsection{Negative Samples}
To model negative captions, we observe the captions' common error types in the model-generated captions. Specifically, we pick 100 bad captions in the order of whose human judgments are low in Composite and Flickr8k, respectively. 
Then, we categorize the main errors into three types:\textit{relevant but have wrong keywords, totally irrelevant to the image, grammatically incorrect.} To model most imperfect captions including these frequent type errors, we prepare negative captions as follows. 
\begin{figure}[t]
\small
\centering
\includegraphics[width=0.90\columnwidth]{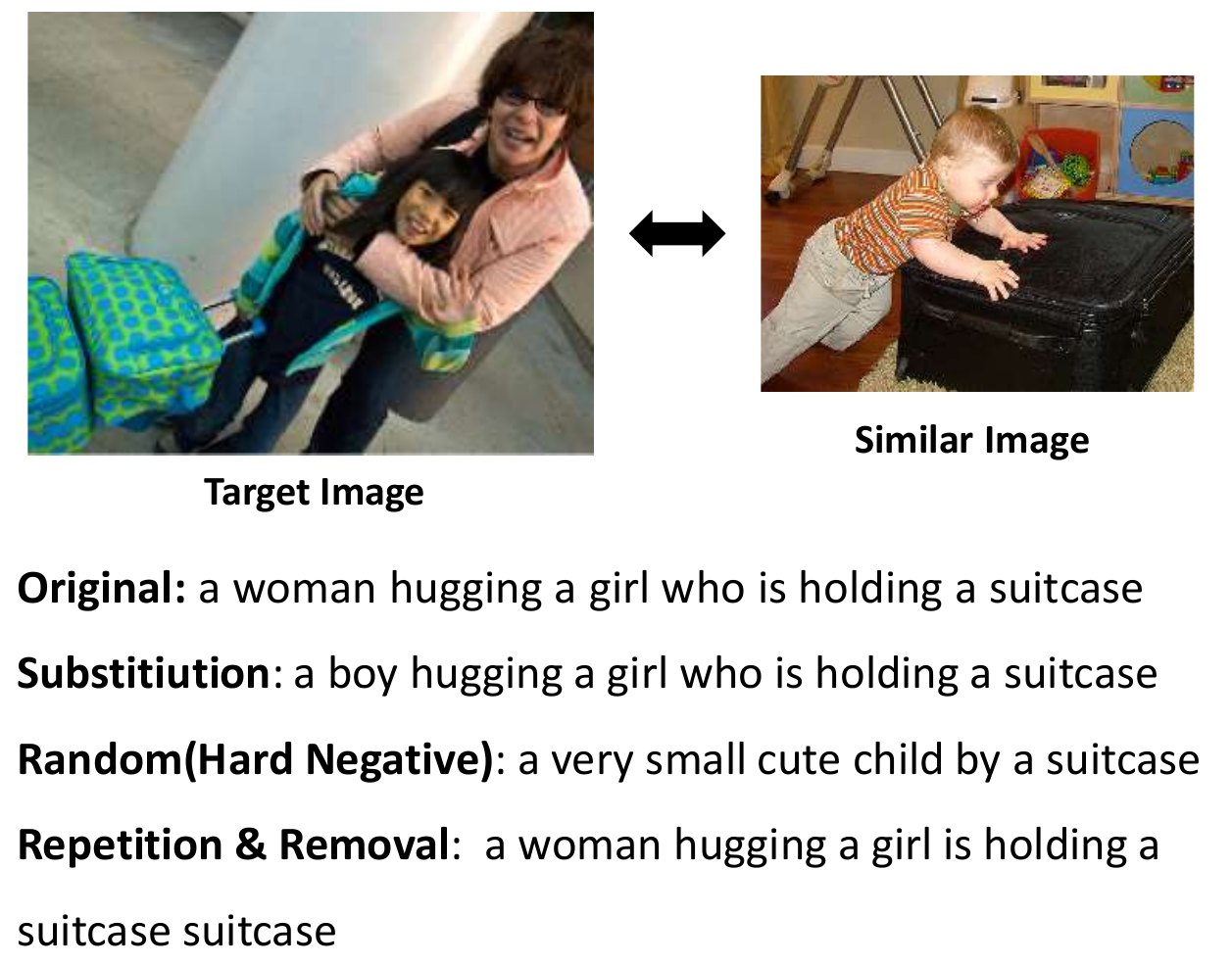}
\caption{
An example of the generated negative captions for the left image to train UMIC. Hard negative caption is one of the reference captions for the right image which is similar to the left image.
}
\vspace{-5mm}
\label{fig_negative}
\end{figure}

\paragraph{Substituting Keywords}
To mimic the captions that are relevant but have wrong keywords, as in the example of Figure~\ref{fig_overall}, we randomly substitute 30\% of the words in the reference captions and use them as negative samples like Figure~\ref{fig_negative}. The motivation we choose 30\% is that the average length of the generated caption is about 10 words and the number of keywords is usually around three. We only substitute \textit{verb}, \textit{adjective}, and \textit{noun}, which are likely to be keywords since they are usually visual words. Also, we substitute them with the words with the same POS-Tags using the pre-defined dictionaries for the captions in the training set to conserve the sentence structure. 

\paragraph{Random Captions}

We randomly sample captions from other images and use them as negative samples to generate totally irrelevant captions for the given image. 
Also, similar to the image-text retrieval task, we use hard-negative captions, which are difficult to be discerned, with a probability of 50\%. Specifically, we utilize the captions of the images similar to the given images using the pre-trained image retrieval model. We get negative captions that are the captions of the similar image sets computed by image-text retrieval model VSE++~\cite{faghri2018vse++} as in~\cite{wang2020compare}. Then, we sample the captions in the reference captions of the Top-3 similar image sets like the example in Figure~\ref{fig_negative}.

\paragraph{Repetition \& Removal}
We find that some of the captions have repeated words or have incomplete sentences. Hence, we randomly repeat or remove some words in the reference captions with a probability of 30\% in the captions to generate these kinds of captions. Specifically, we choose to repeat or remove with a probability of 50\% for the sampled word.

\paragraph{Word Order Permutation}
We further generate negative samples by randomly changing the word order of the reference captions, so that the model sees the overall structure of the sentence, not just the specific visual words.

\subsection{Contrastive Learning}
Using the negative captions generated by the above rules, we fine-tune UNITER via contrastive loss for positive caption $X$ and negative caption $\hat{X}$ as follows.
\begin{equation}
Loss = max(0, M - (\text{S}(I,X) - \text{S}(I,\hat{X}))),
\end{equation}
where $M$ is the margin for the ranking loss, which is a hyperparameter.
We make each batch composed of one positive caption and four negative captions that are made by each negative sample generation technique. 

\section{Dataset}
We briefly explain the previous benchmark datasets for captioning metrics and analyze the problems for two of these datasets, Flickr8k and Composite. Also, we introduce a new benchmark dataset to alleviate the addressed problems.

\subsection{Commonly Used Datasets}
\paragraph{Composite} consists of 11,985 human judgments for each candidate caption generated from three models and image pair. 
This dataset's human judgments range from 1 to 5, depending on the relevance between candidate caption and image.

\paragraph{Flickr8k} provides three expert annotations for each image and candidate caption on 5,822 images. The score ranges from 1 to 4, depending on how well the caption and image match. All of the captions in this dataset are reference captions or captions from other images.

\paragraph{PASCAL50s} contains 1,000 images from UIUC PASCAL Sentence Dataset with 50 reference captions for each image. Different from other datasets, this dataset provides 4,000 caption triplet $<$\textit{A}, \textit{B}, \textit{C}$>$ composed of 50 reference captions(\textit{A}) and two candidate captions(\textit{B}, \textit{C}) for the given image. There are human annotated answers to which is more similar to ``\textit{A}", ``\textit{B}" or ``\textit{C}". 

\subsection{Problems in Flickr8k and Composite}
\label{subsec:prob}
\begin{figure}[t]
\small
\centering
\includegraphics[width=0.95\columnwidth]{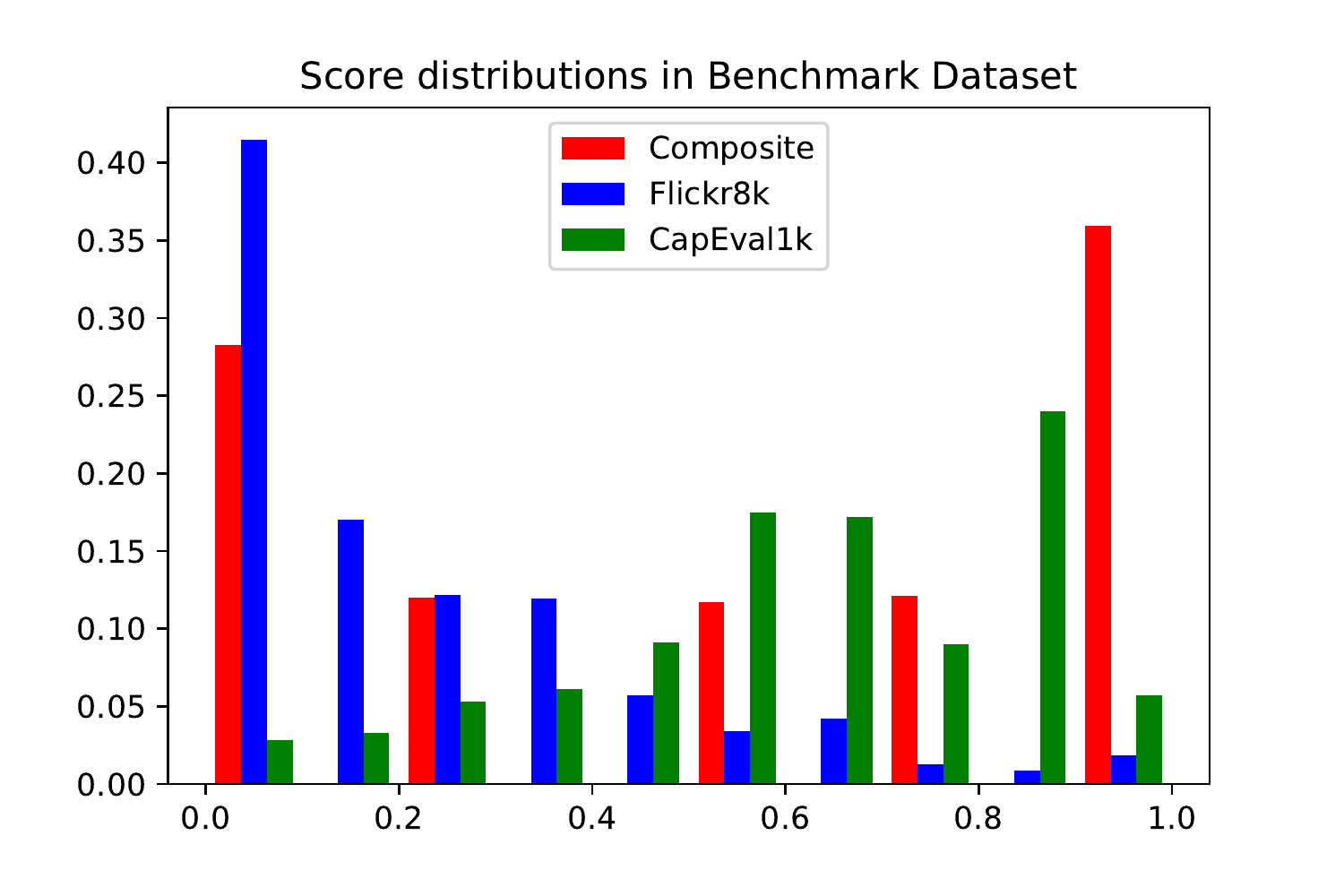}
\caption{
Score distributions of human judgments in Composite, Flickr8k and our proposed CapEval1k dataset. All scores were normalized from 0 to 1.
}
\vspace{-4mm}
\label{fig_score_dist}
\end{figure}
We investigate the human judgments in Flickr8k and Composite, and visualize the distributions of judgment scores for two datasets, Flickr8k and Composite in Figure~\ref{fig_score_dist}, and find several problems.\\
\indent For the Flickr8k, most of the scores are less than 0.2 since the candidate captions were sampled by an image retrieval system from a reference caption pool, not model-generated captions.
Therefore, most captions are not related to images and differ significantly from the model-generated captions. We argue that this naive configuration is not enough to distinguish the performance of the metric precisely.
\\
\indent For the Composite, most of the scores are placed near 0 or 1. We explain this because only a single annotator annotates each sample's score resulting in biased output.
We also manually investigated the captions and found that the captions are coarsely generated. Note that the captions for this dataset were generated by the old model~\cite{karpathy2015deep, aditya2015images}. 
For these reasons, we conclude that additional benchmark dataset is necessary to evaluate the captioning metrics.

\subsection{CapEval1k Dataset}
\label{subsec:capeval1k}
To alleviate the addressed issues in Flickr8k and Composite, we introduce a new dataset CapEval1k, which is composed of human judgments for the model-generated captions from four recently proposed models: Att2in~\cite{rennie2017self}, Transformer~\cite{vaswani2017attention}, BUTD~\cite{anderson2018bottom} and AoANet~\cite{huang2019attention}. Different from Flickr8k and Composite, we ask each annotator to evaluate the captions by considering three dimensions: \textit{fluency}, \textit{relevance}, \textit{descriptiveness}. We hire 5 workers who are fluent in English for each assignment from Amazon Mechanical Turk and use the average score. We also provide the full instructions and details in Appendix.\\
\indent Since our CapEval1k dataset is composed of annotations via recently proposed models, the overall scores are relatively higher than other datasets as shown in Figure~\ref{fig_score_dist}.
Compared to other datasets, CapEval1k contains the annotators' comprehensive judgment across multiple dimensions in evaluating the quality of the generated captions, so we can see that the score distribution score is not concentrated in a particular area.


\section{Experiments}
\subsection{Implementation Details}
We use the pre-trained UNITER-base with 12 layers in the official code provided by the authors~\cite{chen2020uniter}\footnote{https://github.com/ChenRocks/UNITER}. We use the COCO dataset~\cite{fang2015captions} to fine-tune UNITER through ranking loss. We use the train and validation split of COCO dataset in~\cite{chen2020uniter}. The number of the training set is 414k, and the validation set is 25k. We set the batch size of 320, learning rate of 2e-6, and fine-tune UNITER for a maximum of 4k steps. We select the model that shows the minimum loss in the validation set. We set margin $M$ as 0.2 in the ranking loss. We repeat training 5 times for each best-performing model.

\subsection{Performance Comparison}
\begin{table}[!t]
\small
\centering
\resizebox{1\columnwidth}{!}{%
\begin{tabular}
{L{0.23\columnwidth}C{0.16\columnwidth}C{0.22\columnwidth}C{0.24\columnwidth}C{0.24\columnwidth}}
\toprule
\cmidrule{1-5}
           \textbf{Metric} & \textbf{Flickr8k} & \textbf{Composite} & \textbf{CapEval1k} & \textbf{PASCAL50s}\\
\midrule
\textbf{BLEU-1} & 0.274 & 0.406 & 0.233 & 74.3\\
\textbf{BLEU-4} & 0.286 & 0.439 & 0.238 & 73.4\\
\textbf{ROUGE-L} & 0.300 & 0.417 & 0.220 & 74.9\\
\textbf{METEOR} & 0.403 & 0.466 & 0.288 & 78.5\\
\textbf{CIDEr} & 0.419 & 0.473 & 0.307 & 76.1\\
\textbf{SPICE} & 0.457 & 0.486 & 0.279 & 73.6\\
\textbf{BERTScore} & 0.396 & 0.456 & 0.273 & 79.5\\
\textbf{BERT-TBR} & 0.467 & 0.439 & 0.257 & \textbf{80.1}\\
\textbf{VBTScore} & \textbf{0.525} & \textbf{0.514} & \textbf{0.352} & 79.6\\
\midrule
\textbf{VIFIDEL} & 0.336 & 0.191 & 0.143 & 70.0 \\
\textbf{UMIC} & \textbf{0.468} & \textbf{0.561} & \textbf{0.328} & \textbf{85.1}\\
\textbf{UMIC\textsubscript{-C}} & 0.431 & 0.554 & 0.299 & 84.7\\
\bottomrule 
\end{tabular}
}
\caption{
Columns 1 to 3 represent Kendall Correlation between human judgments and various metrics
on Flickr8k, Composite and CapEval1k. All p-values in the results are $<$ 0.01. 
The last column shows the accuracy of matches between human judgments in PASCAL50s. 
}\vspace{-5mm}
\label{table_caplevel}
\end{table}

We compute caption-level Kendall's correlation coefficient with human judgments for the Composite, Flickr8k, and our proposed CapEval1k. For the PASCAL50s, we compute the number of matches between human judgments for each candidate caption pair. For all of the reference based metrics, we use five reference captions and then get average score among the five references except for BERTScore where we use maximum.\\
\indent We present the experimental results for all four datasets in Table~\ref{table_caplevel}.
We show that although UMIC does not utilize any reference captions, UMIC outperforms the baseline metrics except for VBTScore in all of the datasets that depend on multiple references. We also report the strong unreferenced baseline UMIC\textsubscript{-C}, which is directly using the pre-trained weights from UNITER without contrastive learning. Interestingly, UMIC\textsubscript{-C} shows a higher performance than most of the metrics. This high performance shows that pre-trained image-text matching layer of UNITER already has a good representation for evaluating image captions.
Especially for Composite, both UMIC and UMIC\textsubscript{-C} significantly outperform baseline metrics. We explain this in the polarized distribution of human judgments as we explained in Section~\ref{subsec:prob}. In other words, the relevance of most image-caption pairs in this dataset is too obvious so that UNITER can easily distinguish them. However, while UMIC shows higher performance on all datasets, UMIC\textsubscript{-C} shows relatively low performance on Flickr8k and CapEval1k. And this demonstrates the effectiveness and generalization ability of our contrastive learning objective to develop UMIC.

\indent Also, we can observe that the performance of each metric is relatively low and the rank of each metric changes in our proposed CapEval1k dataset. We explain that this is because the captions in CapEval1k are relatively difficult to be evaluated since the score distribution is not biased as explained in Section~\ref{subsec:capeval1k}.

\subsection{Case Study}
\begin{figure}[t]
\small
\centering
\includegraphics[width=1.0\columnwidth]{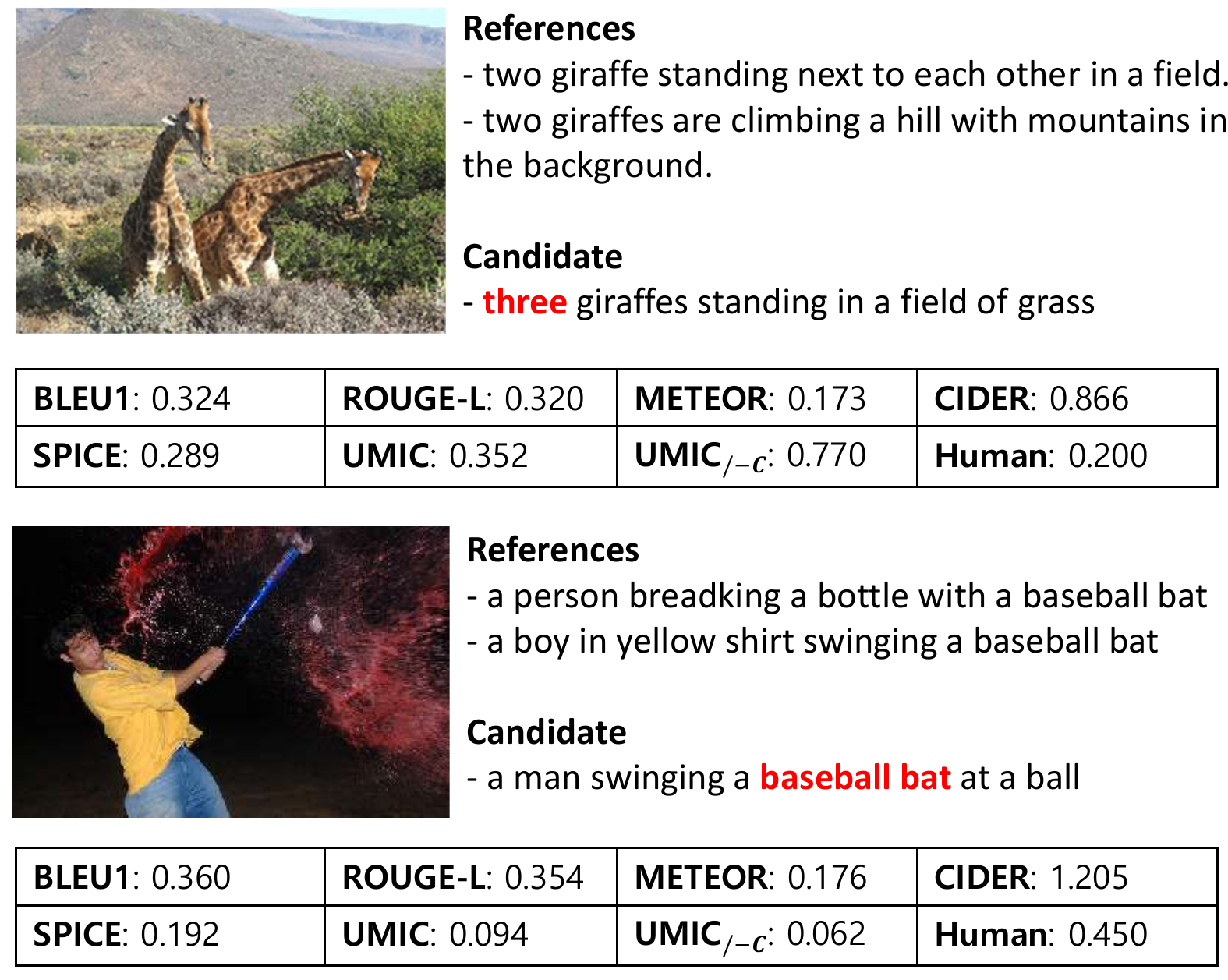}
\caption{
Case study for the various metrics on candidate captions in CapEval1k Dataset. Human judgments are normarlized from 0 to 1.
}
\vspace{-3mm}
\label{fig_case_study}
\end{figure}
We visualize one sample each showing the strengths and weaknesses of UMIC in Figure~\ref{fig_case_study}. 
In the above example, the candidate caption is partially relevant to the image, but the single word ``three" in the caption is totally incorrect since there are only ``two" giraffes in the image. And this leads to a low human judgment of 0.2. Nevertheless, unlike our UMIC, widely used metrics and UMIC\textsubscript{-C} give this caption a high score due to the many words overlaps or missing the keywords. The bottom example shows one of the error cases and the limitations of our proposed method. Since the detection model in UMIC could not recognize the important object like the ``baseball bat", UMIC outputs very low score.

\section{Conclusion}
In this paper, we propose UMIC, an unreferened metric that does not require any reference captions for image captioning task through contrastive learning in UNITER. Also, we propose a new benchmark dataset for image captioning that relieve the issues in previous datasets. 
Experimental results on four benchmark datasets, including our new dataset, show that UMIC outperforms previous metrics.

\section*{Acknowledgements}
We thank anonymous reviewers for their constructive and insightful comments. K. Jung is with ASRI, Seoul National University, Korea. This work was supported by the National Research Foundation of Korea (NRF) grant funded by the Korea government (No. 2021R1A2C2008855). This work was partially funded by gifts from Adobe Research.

\section*{Ethical Considerations}
We compensate the annotators with competitive pay, which is above hourly USD \$10 for collecting human annotated judgments for the model generated captions. Specifically, we pay \$0.2 for each task that is composed of evaluating four candidate captions for a single image, where each task can be usually done in a minute. And we use public datasets to train the models. 

\bibliographystyle{acl_natbib}
\bibliography{acl2021}

\appendix


\section{Experimental Details}

\subsection{Reproducibility Checklist}

\paragraph{Source Code}
We provide the source code for both training UMIC and computing UMIC as supplementary material. We will publicly release the full source with the pre-trained model to easily compute UMIC. 

\paragraph{Computing Infrastructure}
We use AMD Ryzen Threadripper 2950X (3.50 GHz) with GeForce GTX 2080 Ti for the experiments. The software environments are Python 3.6.8 and PyTorch 1.1.0.

\paragraph{Average runtime for each approach}
Each epoch of our training UMIC on average takes 20 minutes using a single GPU. For evaluation, it takes a minute.

\paragraph{Number of Model Parameters} The number of parameters in UMIC is about 109.9M.

\subsection{Correlation Coefficient}
We compute Kendall-C for Flickr8k~\cite{hodosh2013framing}, since we could produce the similar results for most of the previous papers. And we compute Kendall-B for Composite~\cite{aditya2015images} and CapEval1k. For Composite, we use five references and some of the candidate captions in this dataset are exact same with one of the references.

\subsection{Significance Test}
For all of the correlation coefficients we computed in this paper, we conduct a standard way to test the significance of the correlation coefficient. We use a t-test using a null hypothesis that is an absence of association to report the p-value for each coefficient.

\section{Data Collection}
\subsection{Generating Captions}
We generate the captions from the images in Karphathy's test split that do not have any overlaps in the training set and validation set of UMIC.
We use four models, Att2in~\cite{rennie2017self}, Transformer~\cite{vaswani2017attention}, BUTD~\cite{anderson2018bottom}, and AoANet~\cite{huang2019attention} to generate captions.
We use the pre-trained model that uses self-critical loss~\cite{luo2018discriminability} in the public repository~\footnote{https://github.com/ruotianluo/self-critical.pytorch}. We set beam size 2 for all of the models during the inference. We sample 1,000 captions for a total of 250 images for each model, where each caption does not have a single equivalent as shown in Figure~\ref{fig_instructions}.

\subsection{Instructions to Annotators}
\begin{figure*}[t]
\small
\centering
\includegraphics[width=1.6\columnwidth]{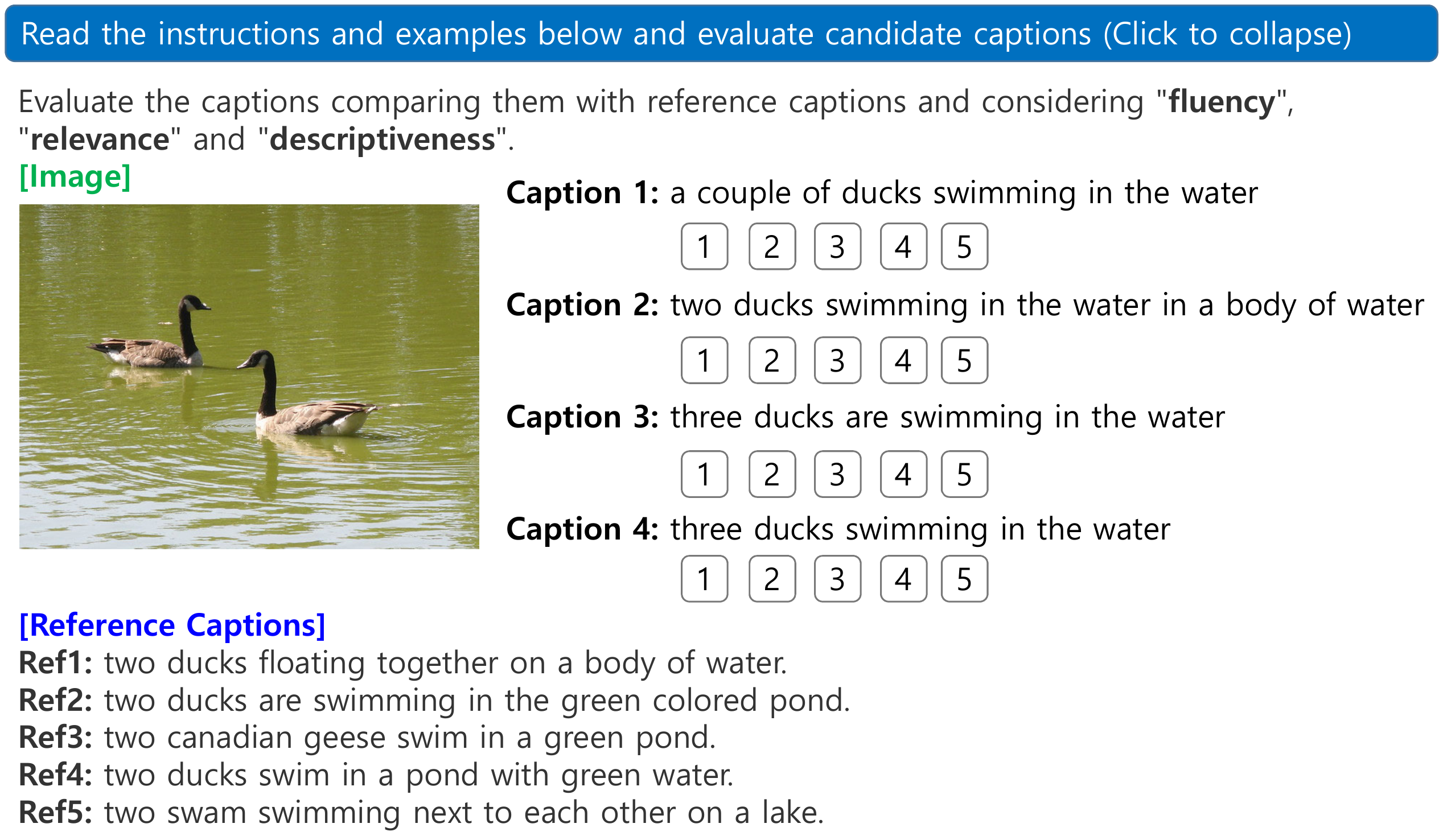}
\caption{
Annotation interface and short instructions for captioning evaluation task.
}
\label{fig_instructions}
\end{figure*}

\begin{figure}[t]
\small
\centering
\includegraphics[width=0.9\columnwidth]{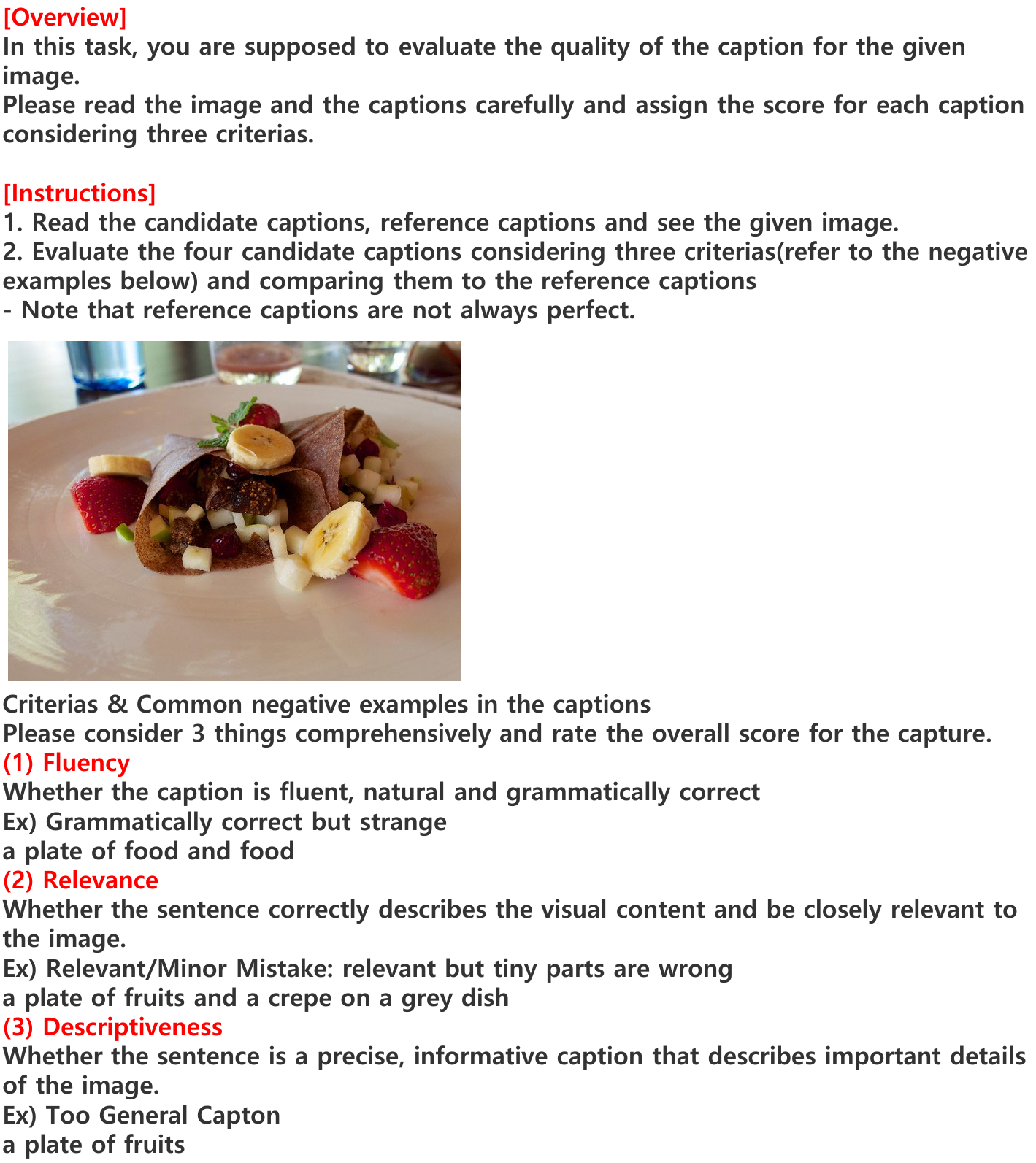}
\caption{
Full instructions for the captioning evaluation task. We provide an image and five reference captions to the workers and request them to evaluate four captions.
}
\label{fig_full_instructions}
\end{figure}
The interface and instructions to annotators in MTurk are shown in Figure~\ref{fig_instructions} and Figure~\ref{fig_full_instructions}. We request the worker to evaluate four captions at once in a single assignment so that the worker can consider the difference among the captions.

\subsection{Inter-annotator Agreement} We compute the annotator agreement using Krippendorff's $\alpha$~\cite{krippendorff1970estimating}. We observe that Krippendorff's $\alpha$ is 0.37 that indicates a ``fair`` agreement according to one of the general guidelines~\cite{landis1977measurement} for kappa-like measures.

\subsection{Worker Pool \& Pay}
We hire the annotators whose locations in one of the US, UK, CA, NZ, AU. We restrict the workers whose HIT approval rates are higher than 96\%, and minimum hits are over 5000. We pay workers more than USD \$10 in an hour through several preliminary experiments on the compensation.

\end{document}